\documentclass{article}
\usepackage[final]{nips_2018}

\usepackage[utf8]{inputenc} % allow utf-8 input
\usepackage[T1]{fontenc}    % use 8-bit T1 fonts
\usepackage{hyperref}       % hyperlinks
\usepackage{url}            % simple URL typesetting
\usepackage{booktabs}       % professional-quality tables
\usepackage{amsfonts}       % blackboard math symbols
\usepackage{nicefrac}       % compact symbols for 1/2, etc.
\usepackage{microtype}      % microtypography
\usepackage{placeins}

\usepackage{times}
\usepackage{graphicx} % more modern
\usepackage{subfigure} 

\usepackage{rotating}

\usepackage{natbib}

\usepackage{algorithm}
\usepackage{algpseudocode} 

\usepackage{pgfplots}

\usepackage{amsmath,amssymb}

\title{Learning Montezuma's Revenge\\ from a Single Demonstration}

\author{
Tim Salimans\\
OpenAI, Google Brain
\And
Richard Chen\\
OpenAI, Happy Elements Inc.
}

\begin{document} 

\maketitle

\begin{abstract}
We propose a new method for learning from a single demonstration to solve hard exploration tasks like the Atari game Montezuma's Revenge. Instead of imitating human demonstrations, as proposed in other recent works, our approach is to maximize rewards directly. Our agent is trained using off-the-shelf reinforcement learning, but starts every episode by resetting to a state from a demonstration. By starting from such demonstration states, the agent requires much less exploration to learn a game compared to when it starts from the beginning of the game at every episode. We analyze reinforcement learning for tasks with sparse rewards in a simple toy environment, where we show that the run-time of standard RL methods scales exponentially in the number of states between rewards. Our method reduces this to quadratic scaling, opening up many tasks that were previously infeasible. We then apply our method to Montezuma's Revenge, for which we present a trained agent achieving a high-score of 74,500, better than any previously published result.

%We have trained an agent from a single human demonstration to learn to achieve a final score of 74,500 on Montezuma’s Revenge, better than any previously published result. Our algorithm is simple: the agent learns via reinforcement learning, but starts every episode by resetting to a state from a demonstration. By starting from demonstration states, the agent needs to perform much less exploration to learn to play the game compared to when it starts from the beginning of the game at every episode. Contrary to other recent approaches to learning Montezuma's Revenge from demonstrations, we do not train the agent to imitate the demonstration; the agent is trained to maximize the score using pure RL.
\end{abstract}

\section{Introduction}
Model-free reinforcement learning is learning by trial and error. Methods such as policy gradients (\cite{williams1992simple, sutton2000policy, kakade2002natural}) and Q-learning (\cite{watkins1992q, mnih2015human}) explore an environment by taking random actions. If, by chance, the random actions lead to success, formulated as achieving a \text{reward}, they are \emph{reinforced} and the agent becomes more likely to take these beneficial actions again in the future. This works well if rewards are frequent enough for random actions to lead to a reward with reasonable probability. Unfortunately, many of the tasks we would like to solve don't have such dense rewards, instead requiring long sequences of very specific actions to achieve any success. Such sequences are extremely unlikely to occur randomly.

Consider a task where it takes a precise sequence of $N$ actions to achieve the first reward. If each of those actions is taken with a fixed probability, a random agent will need to explore this environment for a duration that scales as $\exp(N)$ before it can expect to experience the first reward. A well-known example of such a task is the Atari game Montezuma's Revenge, where the goal is to navigate a series of chambers to collect keys, diamonds, and other items, while evading various opponents and traps. In this game, the probability of achieving the first reward of the game, the key in the first room, can be decomposed as
\begin{multline*}
   p(\text{get first key}) = p(\text{get down ladder 1}) * p(\text{get down rope}) * \\
    p(\text{get down ladder 2}) * p(\text{jump over skull}) * p(\text{get up ladder 3}). 
\end{multline*}
By multiplying $N$ of these probabilities together, we end up with the resulting probability $p(\text{get key})$ that is exponentially smaller than any of the individual input probabilities. As a result, taking uniformly random actions in Montezuma's Revenge only produces a reward about once in every half a million steps. This exponential scaling of reinforcement learning severely limits the tasks current RL techniques can solve.

Prior works (\cite{dearden1999model, kolter2009near, Tang2016exploration, ostrovski2017count, chen2017ucb, pathak2017curiosity, o2016pgq, schulman2017equivalence, haarnoja2017reinforcement, nachum2017bridging}) have proposed various methods of overcoming the exploration problems of naive model-free RL. However, these methods have so far produced only limited gains for solving hard exploration tasks such as Montezuma's Revenge. In this work, we instead consider using a single \emph{demonstration} of successful completion of the task to aid exploration in reinforcement learning. Most previous work on learning from demonstrations has so far focused on \emph{imitation learning}, where the agent is trained to imitate the demonstration. A downside of this approach is that many different demonstrations are required for the agent to learn to generalize, and that these demonstrations need to be of high quality to avoid learning a sub-optimal solution. Here, we instead show that it is feasible to learn to solve sparse reward problems like Montezuma's Revenge purely by RL, by bypassing the exploration problem through starting each episode from a carefully selected state from the demonstration.

\nocite{bellemare2013arcade}
\section{Method}
Although model-free RL methods have difficulty finding long sequences of actions, they work well for shorter sequences. The main insight behind our proposed method is that we can make a task easier to solve by decomposing it into a curriculum of subtasks requiring short action sequences. We construct this curriculum by starting each RL episode from a demonstration state. We implement this idea in a distributed setting, using $M$ parallel \emph{rollout workers} that collect data by acting in an environment according to a shared RNN policy $\pi(\theta)$, after which the data is fed to a centralized \emph{optimizer} to learn an improvement of the policy. 

Given a previously recorded demonstration $\{(\tilde{s}_t, \tilde{a}_t, \tilde{r}_t, \tilde{s}_{t+1})\}_{t=0}^T$, our approach works by letting each parallel rollout worker (Algorithm~\ref{algo:rollout}) start its episode from a state $\tilde{s}_{\tau^{*}}$ in the demonstration. Early on in training, all workers start from states at times $\tau^{*}$ near the end of the demonstration at time $T$. These reset points are then gradually moved back in time as training proceeds. The data produced by the rollout workers is fed to a central optimizer (Algorithm~\ref{algo:optim}) that updates the policy $\pi(\theta)$ using an off-the-shelf RL method such as PPO (\cite{schulman2017proximal}), A3C (\cite{mnih2016asynchronous}), or Impala (\cite{espeholt2018impala}). In addition, the central optimizer calculates the proportion of rollouts that beat or at least tie the score of the demonstrator on the corresponding part of the game. If the proportion is higher than a  threshold $\rho$, we move the reset point $\tau$ backward in the demonstration. 

Within each iteration of Algorithm~\ref{algo:rollout}, the rollout workers obtain the latest policy $\pi(\theta)$ and the central reset point $\tau$ from the optimizer. Each worker then samples a local starting point $\tau^*$ from a small set of time steps $\{\tau - D, \dots, \tau\}$ to increase diversity. In each episode, we first initialize the agent's RNN policy's hidden states by taking $K$ actions based on the demonstration segment $\{(\tilde{s}_t, \tilde{a}_t, \tilde{r}_i, \tilde{s}_{i+1})\}_{i=\tau^*-K}^{\tau^*-1}$ directly preceding the local starting point $\tau^*$, after which the agent takes actions based on the current policy $\pi(\theta)$. The demonstration segment used for initializing the RNN states is masked out in the data $\mathcal{D}$ used for training, such that it does not contribute to the gradient used in the policy update. At the end of each episode, we increment a success counter $W$ if the current episode achieved a high score compared to the score in demonstration, which is then used to decrease the central starting point $\tau$ at the right speed.

Training proceeds until the central reset point has reached the beginning of the game, i.e.\ $\tau=0$, so that the agent is succeeding at the game without using the demonstration at all. At this point we have an RL-trained agent beating or tying the human expert demonstration on the entire game.

\begin{algorithm}
\caption{Demonstration-Initialized Rollout Worker}
\label{algo:rollout}
\begin{algorithmic}[1]
\State {\bf Input:} a human demonstration $\{(\tilde{s}_t, \tilde{a}_t, \tilde{r}_t, \tilde{s}_{t+1}, \tilde{d}_t)\}_{t=0}^T$, number of starting points $D$, effective RNN memory length $K$, batch rollout length $L$.
\State Initialize starting point $\tau^{*}$ by sampling uniformly from $\{\tau - D, \dots, \tau\}$
\State Initialize environment to demonstration state $\tilde{s}_{\tau^{*}}$
\State Initialize time counter $i = \tau^{*}-K$
\While{\text{TRUE}}
    \State Get latest policy $\pi(\theta)$ from optimizer
    \State Get latest reset point $\tau$ from optimizer
    \State Initialize success counter $W = 0$
    \State Initialize batch $\mathcal{D}=\{\}$
    \For{step in $0,\ldots,L-1$}
        \If{$i \geq \tau^{*}$}
            \State Sample action $a_{i} \sim \pi(s_{i}, \theta)$
            \State Take action $a_{i}$ in the environment
            \State Receive reward $r_{i}$, next state $s_{i+1}$ and done signal $d_{i+1}$
            \State $m_{i}=\text{TRUE}$ \Comment{We can train on this data}
        \Else \Comment{Replay demonstration to initialize RNN state of policy}
            \State Copy data from demonstration $a_{i}=\tilde{a}_i, r_{i}=\tilde{r}_i, s_{i+1}=\tilde{s}_{i+1}, d_{i}=\tilde{d}_{i}$.
            \State $m_{i}=\text{FALSE}$ \Comment{We should mask out this transition in training}
        \EndIf
        \State Add data $\{s_{i},a_{i},r_{i},s_{i+1},d_{i},m_{i}\}$ to batch $\mathcal{D}$
        \State Increment time counter $i \leftarrow i+1$
        \If{$d_{i} = $ TRUE}
            \If{$\sum_{t \geq \tau^{*}}r_{t} \geq \sum_{t \geq \tau^{*}} \tilde{r}_t$} \Comment{As good as demo}
                \State $W \leftarrow W+1$
            \EndIf
            \State Sample next starting starting point $\tau^{*}$ uniformly from $\{\tau-D, ...,\tau\}$
            \State Set time counter $i \leftarrow \tau^{*}-K$
            \State Reset environment to state $\tilde{s}_{\tau^{*}}$
        \EndIf
    \EndFor
    \State Send batch $\mathcal{D}$ and counter $W$ to optimizer 
\EndWhile
\end{algorithmic}
\end{algorithm}

\begin{algorithm}
\caption{Optimizer}
\label{algo:optim}
\begin{algorithmic}[1]
\State {\bf Input:} number of parallel agents $M$, starting point shift size $\Delta$, success threshold $\rho$, initial parameters $\theta_{0}$, demonstration length $T$, learning algorithm $\mathcal{A}$ (e.g.\ PPO, A3C, Impala, etc.)
\State Set the reset point $\tau = T$ to the end of the demonstration
\State Start rollout workers $i=0,\ldots,M-1$
\While{$\tau > 0$}
    \State Gather data $\mathcal{D} = \{\mathcal{D}_{0},\ldots,\mathcal{D}_{M-1}\}$ from rollout workers
    \If{$\sum_{\mathcal{D}}[W_{i}]/\sum_{\mathcal{D}}[d_{i,t}] \geq \rho$} \Comment{The workers are successful sufficiently often}
        \State $\tau \leftarrow \tau - \Delta$
    \EndIf
    \State $\theta \leftarrow \mathcal{A}(\theta,\mathcal{D})$ \Comment{Make sure to mask out demo transitions}
    \State Broadcast $\theta, \tau$ to rollout workers
\EndWhile
\end{algorithmic}
\end{algorithm}

By slowly moving the starting state from the end of the demonstration to the beginning, we ensure that at every point the agent faces an easy exploration problem where it is likely to succeed, since it has already learned to solve most of the remaining game. We can interpret solving the RL problem in this way as a form of dynamic programming \citep{bagnell2004policy}. If a specific sequence of $N$ actions is required to reach a reward, this sequence may now be learned in a time that is quadratic in $N$, rather than exponential. Figure~\ref{fig:montezuma} demonstrates this intuition in Montezuma's Revenge.

\begin{figure}[h]
    \centering
    \includegraphics[width=\textwidth]{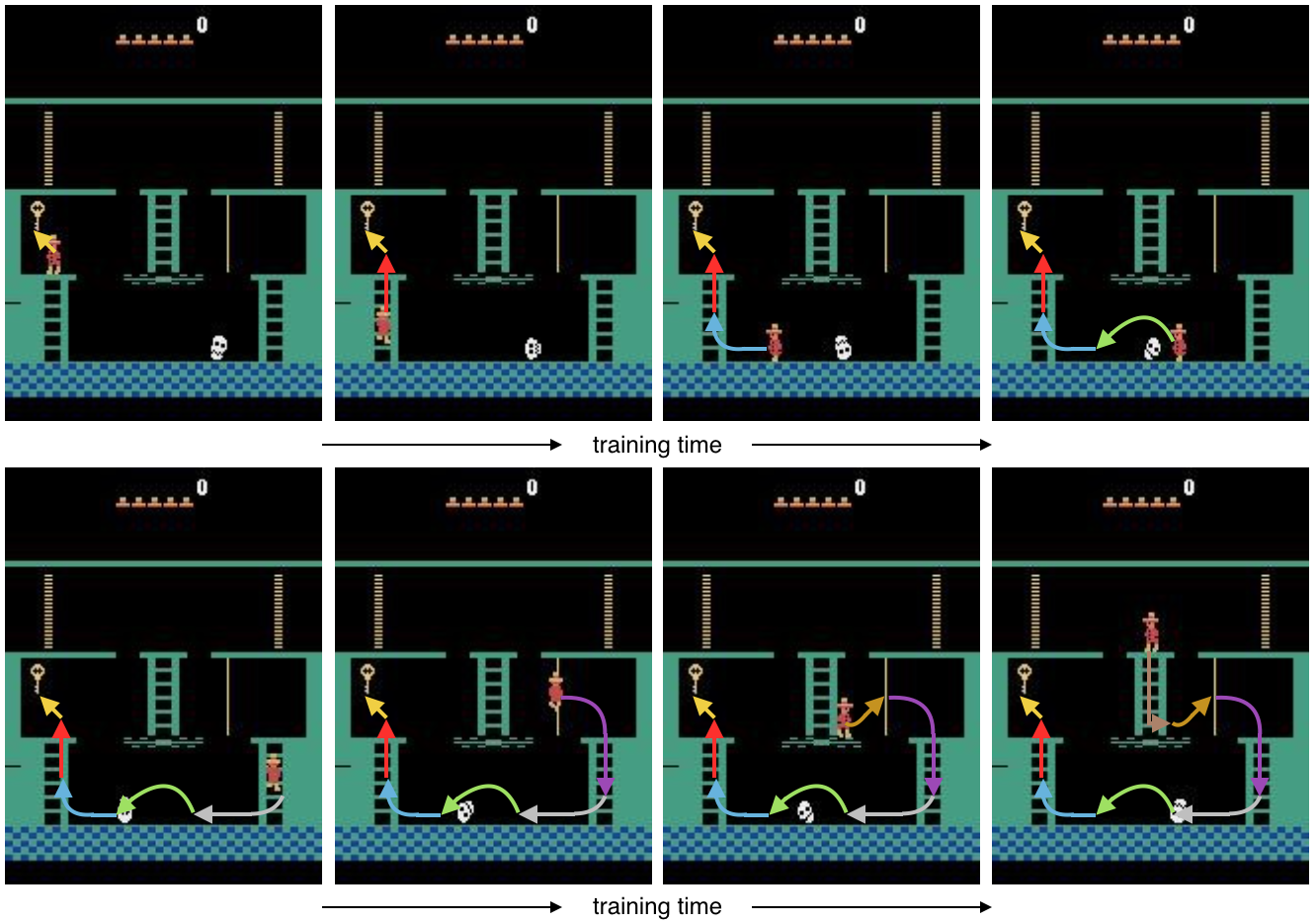}
    \caption{Impression of our agent learning to reach the first key in Montezuma’s Revenge using RL and starting each episode from a demonstration state. When our agent starts playing the game, we place it right in front of the key, requiring it to only take a single jump to find success. After our agent has learned to do this consistently, we slowly move the starting point back in time. Our agent might then find itself halfway up the ladder that leads to the key. Once it learns to climb the ladder from there, we can have it start at the point where it needs to jump over the skull. After it learns to do that, we can have it start on the rope leading to the floor of the room, etc. Eventually, the agent starts in the original starting state of the game and is able to reach the key completely by itself.}
    \label{fig:montezuma}
\end{figure}
\FloatBarrier

\section{Related work}
\subsection{State resetting}
Starting episodes by resetting from demonstration states was previously proposed by \cite{hosu2016playing} for learning difficult Atari games. However, their method did not construct a curriculum that gradually moves the starting state back from the end of the demonstration to the beginning. We found such a curriculum to be vitally important for deriving any benefit from the demonstration. 

The idea of constructing a learning curriculum by starting each RL episode from a sequence of increasingly more difficult starting points was used recently by \cite{florensa2017reverse} for an application in robotics. Rather than selecting states from a demonstration, the authors construct the curriculum by iteratively perturbing a set of starting states using random actions and then selecting the resulting states with the right level of difficulty.

After the release of an early version of this work, \cite{resnick2018backplay} published concurrent work exploring the construction of a curriculum from demonstration states, showing promising results in both single-agent and multi-agent tasks. Their curriculum also starts with states near the end of the game and moves gradually to the beginning. In contrast to our dynamically learned curriculum, theirs is predefined for each task. We also explored this in our experiments but found the dynamic adjustment of the starting state to be crucial to achieve good results on difficult tasks like Montezuma's Revenge.

\subsection{Imitation Learning}
Another relevant research direction aims to solve the exploration problem via imitation of a human expert. One example is the work by \cite{peng2018deepmimic} which uses imitation learning to mimic demonstrated movements for use in physics-based animation. Another example is the work by \cite{nair2018overcoming} that combines demonstration-based imitation learning and reinforcement learning to overcome exploration problems for robotic tasks.

Recently, several researchers successfully demonstrated an agent learning Montezuma's Revenge by imitation learning from a demonstration. \cite{aytar2018playing} train an agent to achieve the same states seen in a YouTube video of Montezuma's Revenge, where \cite{pohlen2018observe} combine a sophisticated version of Q-learning with maximizing the likelihood of actions taken in a demonstration. \cite{garmulewicz2018expert} proposed the expert-augmented actor critic method, which combines the ACKTR policy gradient optimizer (\cite{wu2017scalable}) with an extra loss term based on supervision from expert demonstrations, also obtaining strong results on Montezuma's Revenge.

The advantage of approaches based on imitation is that they do not generally require as much control over the environment as our technique does: they do not reset the environment to states other than the starting state of the game, and they do not presume access to the full game states encountered in the demonstration. Our method differs by directly optimizing what we care about — the game score, rather than making the agent imitate the demonstration; our method can thus learn from a single demonstration without overfitting, is more robust to potentially sub-optimal demonstrations, and could offer benefits in multi-agent settings where we want to optimize performance against other opponents than the ones seen in the demonstration.

\section{Experiments}
We test our method on two environments. The first is the blind cliff walk environment of \cite{schaul2015prioritized}, where we demonstrate that our method reduces the exponential exploration complexity of conventional RL methods to quadratic complexity. The second is the notoriously hard Atari game Montezuma's Revenge, where we achieve a higher score than any previously published result. We discuss results, implementation details, and remaining challenges. 

\subsection{Blind cliff walk} \label{section:cliff}
To gain insight into our proposed algorithm we start with the blind cliff walk environment proposed by \cite{schaul2015prioritized}. This is a simple RL toy problem where the goal is for the agent to blindly navigate a one dimensional cliff. The agent starts in state 0 and has 2 available actions. One of these actions will take it to the next state, while the other action will make it fall off the cliff, at which point it needs to start over. Only when the end of the cliff is reached (the last state of $N$ states) does the agent receive a reward. We assume there is no way for the agent to generalize across the different states, so that the agent has to learn a tabular policy.

\begin{figure}[H]
    \centering
    \includegraphics[width=0.7\textwidth]{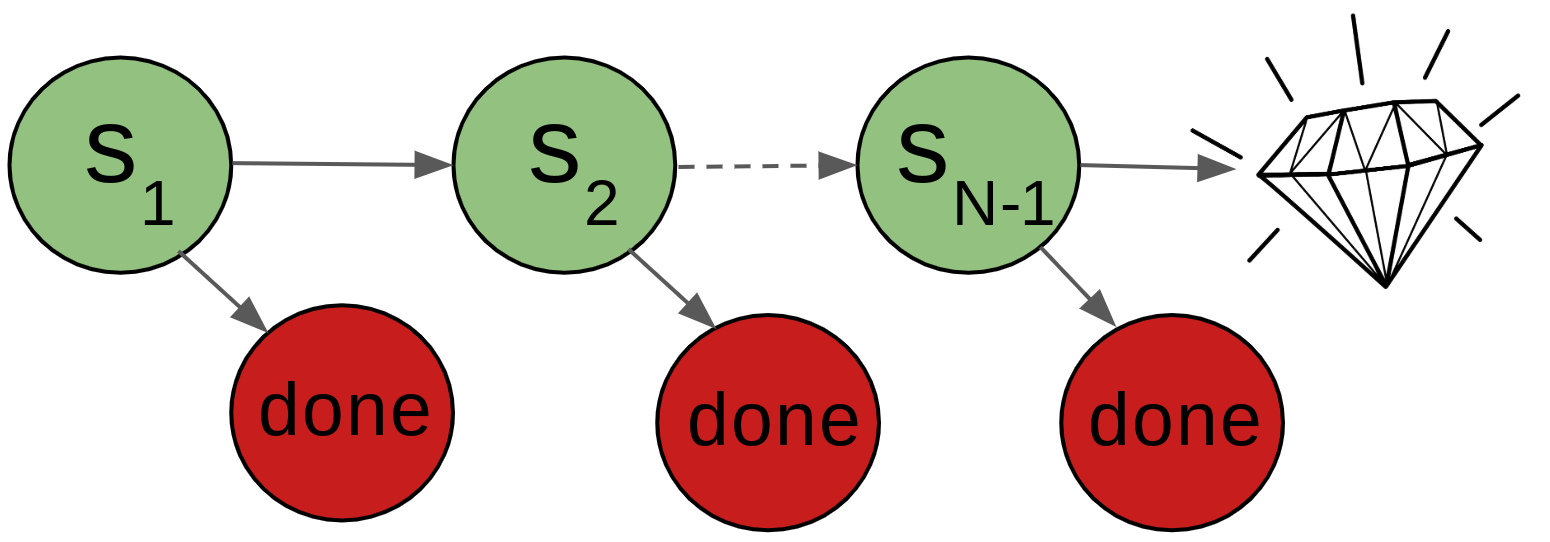}
    \caption{Visualization of the blind cliff walk problem. A single correct sequence of actions traverses all states, leading to a reward. The agent learns a tabular policy for this problem and does not generalize between states. We can vary the problem size $N$ to make the problem more or less difficult.}
    \label{fig:montezuma}
\end{figure}

\begin{figure}[H]
    \centering
    \includegraphics[width=0.7\textwidth]{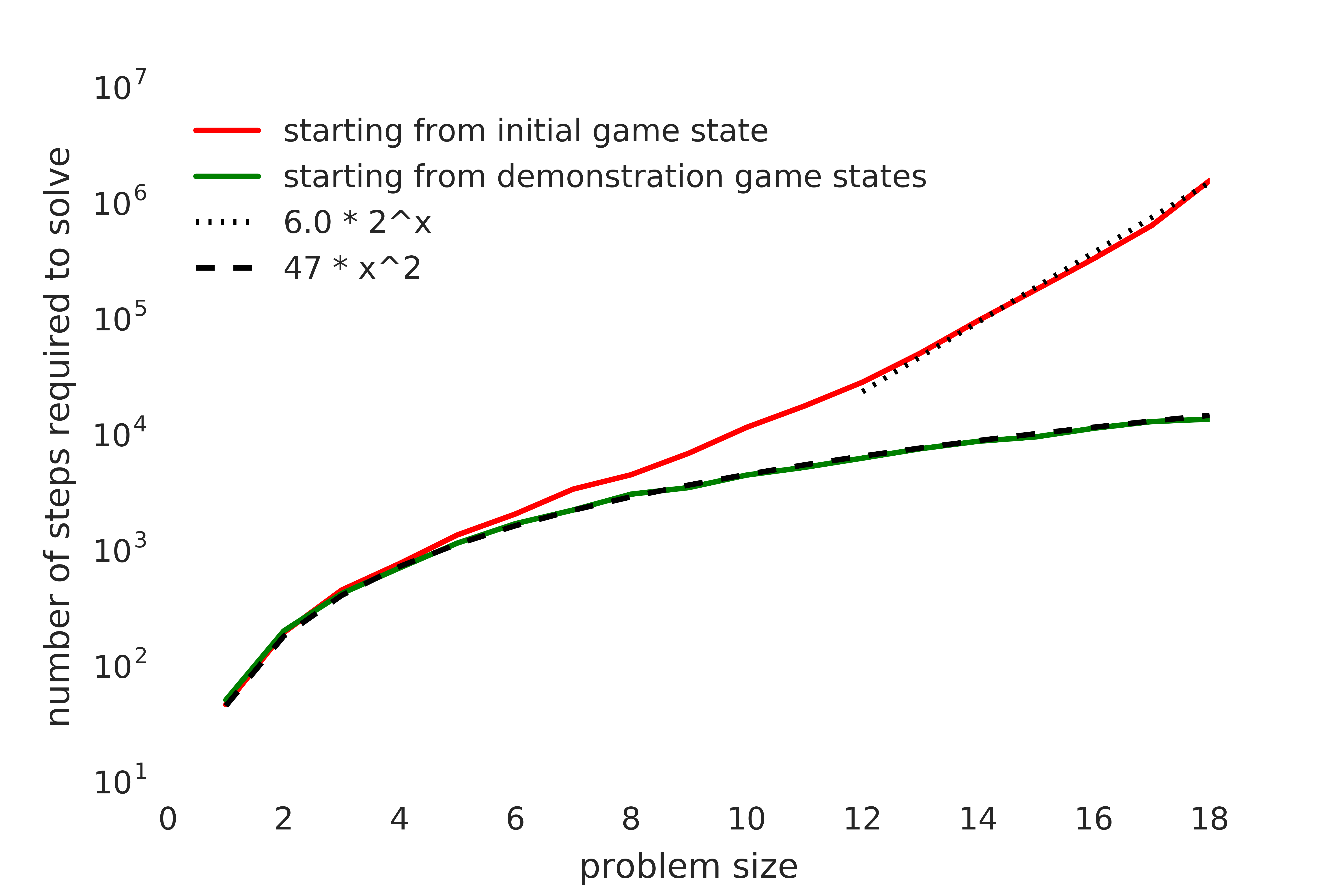}
    \caption{Number of steps required by a standard policy gradients optimizer to learn a policy that solves the blind cliff walk problem with $95\%$ probability. We compare starting each episode from the initial game state versus starting each episode by selecting a state from the demonstration. The reported numbers are geometric means taken over 20 different random seeds. When starting from demonstration states selected using our proposed algorithm, the run time of the optimizer scales quadratically in the problem size. When starting each episode from the initial game state, as is standard practice, the run time scales exponentially in the problem size.}
    \label{fig:cliffwalk}
\end{figure}

As explained by \cite{schaul2015prioritized}, naive application of RL to this problem suffers a run time than scales exponentially in the problem size $N$. The reason is that takes the agent on the order of $2^{N}$ random steps to achieve a reward so that it can learn. Fortunately we can do better when we're given a demonstration of an agent successfully completing the cliff walk. We start by having the agent start each episode at state $N-1$, the second to last state from the demonstration. This means it only needs to take a single right action to receive a reward, so learning is instant. After learning has been successful from state $N-2$, we can start our episodes at state $N-3$, etc. This is expected to give a total run time that scales quadratically in the problem size $N$, as the agent needs to take on the order of $N$ steps to learn what to do in each of the $N$ states. This is a huge improvement over the exponential scaling of the naive RL implementation. Empirically, Figure \ref{fig:cliffwalk} shows that this advantage in scaling indeed holds in practice for this simple example.

\FloatBarrier

\subsection{Montezuma's Revenge} \label{section:montezuma}

\paragraph{Demonstration}
We provide a single demonstration to our algorithm that we recorded by playing the game tool-assisted, using a tool that allowed us to reverse game time and correct any mistakes we made. We have open-sourced the tool we built for this purpose. The score obtained in our demonstration is $71,500$, corresponding to about $15$ minutes of playing time. Although the tool was necessary for us to reach this score as novice players, it is still more than a factor of 10 below the high scores reported by expert players on this game without using external tools.

\paragraph{Implementation}
Our agent for playing Montezuma's Revenge is parameterized by a convolutional neural network, combining standard spatial convolutions with causal convolutions in the time dimension as used in Wavenet \citep{van2016wavenet}. The agent receives grayscale observations of size $105 \times 80$ which are first passed through a 2D convolutional layer with a kernel size of $7\times5$, stride of $2\times2$, and $64$ channel output. Subsequently we apply $4$ 3D convolutional layers where the kernel size in the spatial dimension is $5\times5$, $5\times5$, $5\times5$ and $5\times3$. Each of these layers has a kernel size of 2 in the time dimension, with stride increasing exponentially as $2^{l}$ for layer $l$. The number of channels is doubled at every layer. The resulting network is significantly larger than what is commonly used for reinforcement learning in the Atari environment, which we found to be necessary to learn to beat our lengthy demonstration.

We choose the reset point adjustment threshold used in Algorithm~\ref{algo:optim} as $\rho = 20\%$, such that we move the episode starting point back in time if at least $20\%$ of the rollout workers achieve returns comparable to the provided demonstration. We use PPO \citep{schulman2017proximal} to train the agent's policy, and distribute learning over $128$ GPUs with 8 workers each, for a total of $M=1024$ rollout workers. The agent was trained for about $50$ billion frames.

\paragraph{Result}
Our trained agent achieves a final score of $74,500$ over approximately $12$ minutes of play, a double-speed recording of which is available at \url{https://tinyurl.com/ybbheo86}. We observe that although much of the agent’s game mirrors our demonstration, the agent surpasses the demonstration score of $71,500$ by picking up more diamonds along the way. In addition, the agent makes use of a feature of the game that was unknown to us: at minute 4:25 of the video sufficient time has passed for a key to re-appear, allowing the agent to proceed in a different way from the demonstration.

Table \ref{tab:score} compares our obtained score to results previously reported in the literature. Unfortunately, there is no standard way of evaluating performance in this setting: the game is deterministic, and different methods add different amounts of noise during action selection. Our result was achieved by sampling from a trained policy with low entropy. The amount of noise added is thus small, but comparable to the previous best results such as those by \cite{pohlen2018observe} and \cite{aytar2018playing}.

\begin{table}[h]
\begin{center}
\begin{tabular}{lcc}
\hline
    Approach  &  Score \\
\hline
Count-based exploration (\cite{ostrovski2017count}) & 3,705.5 \\
Unifying count-based exploration (\cite{bellemare2016unifying} ) & 6,600 \\
DQfD (\cite{hester2017deep}) & 4,739.6 \\
 Ape-X DQfD (\cite{pohlen2018observe}) & 29,384\\
Playing by watching Youtube (\cite{aytar2018playing}) & 41,098  \\
\hline 
   Ours  &  74,500 \\
\hline
\end{tabular}
\caption{Score comparison on Montezuma's Revenge}
\label{tab:score}
\end{center}
\end{table}

Like is often the case in reinforcement learning, we find that our trained neural net policy does not yet generalize at the level of a human player. One method to test for generalization ability proposed by \cite{machado2017revisiting} is to perturb the policy by making actions \emph{sticky} and repeating the last action with probability of 0.25 at every frame. Using this evaluation method our trained policy obtains a score of 10,000 on Montezuma's Revenge on average. Alternatively, we can take random actions with probability $0.01$ (repeated for $4$ frameskipped steps), which leads to an average score of $8,400$ for our policy. Anecdotally, we find that such perturbations also significantly reduce the score of human players on Montezuma's Revenge, but to a lesser extent. As far as we are aware, our results using perturbed policies are still better than all those published previously. Perturbing the learned policy by starting with between 0 and 30 random no-ops did not significantly hurt results, with the majority of rollouts achieving at least the final score obtained in our demonstration.

Training our agent to the reported result required 128 GPUs over a period of 2 weeks, which unfortunately made it impossible to quantify the consistency of our algorithms across runs. We leave a more systematic study of the reliability of the algorithm for future work.

\paragraph{Remaining Challenges}
Although the step-by-step learning done by our agent is much simpler than learning to play from scratch, it is still far from trivial. One challenge our RL agent faces is that it is generally unable to reach the exact state from later on in a demonstration when it starts from an earlier state. This is because the agent plays the game at a different frameskip from what we used for recording the demonstration, but it is also due to the randomness in the actions which make it very unlikely to exactly reproduce any specific sequence of actions. The agent will thus need to be able to generalize between states that are very similar, but not identical. We found that this works well for Montezuma’s Revenge, but much less well for some other Atari games we tried, like Gravitar and Pitfall. One reason for this may be that these latter games require solving a harder vision problem: we found these games difficult to play from a downsampled screen ourselves, and we saw some improvement when using larger and deeper neural network policies.

Another challenge we encountered is that standard RL algorithms like policy gradients require striking a careful balance between exploration and exploitation: if the agent’s actions are too random, it makes too many mistakes to ever achieve the required final score when starting from the beginning of the game; if the actions are too deterministic, the agent stops learning because it does not explore alternative actions. Achieving the reported result on Montezuma’s Revenge thus required careful tuning of the coefficient of the entropy bonus used in PPO, in combination with other hyperparameters such as the learning rate and the scaling of rewards. For some other games like Gravitar and Pitfall we were unable to find hyperparameters that worked for training the full curriculum. We hope that future advances in RL will yield algorithms that are more robust to random noise and to the choice of hyperparameters.

\section{Conclusion}
Prior work on learning from demonstrations to solve difficult reinforcement learning tasks has focused mainly on imitation, which encourages identical behavior to that seen in the demonstration. In contrast, we propose a new method that optimizes returns directly. Our method breaks down a difficult exploration problem into a curriculum of subtasks, created by resetting from demonstration states. Our agent does not mimic the demonstrated behavior exactly and is able to find new and exciting solutions that the human demonstrator may not have considered, resulting in a higher score on Montezuma's Revenge than obtained using previously published approaches.

\bibliography{bibliography}

\begin{thebibliography}{34}
\providecommand{\natexlab}[1]{#1}
\providecommand{\url}[1]{\texttt{#1}}
\expandafter\ifx\csname urlstyle\endcsname\relax
  \providecommand{\doi}[1]{doi: #1}\else
  \providecommand{\doi}{doi: \begingroup \urlstyle{rm}\Url}\fi

\bibitem[Aytar et~al.(2018)Aytar, Pfaff, Budden, Paine, Wang, and
  de~Freitas]{aytar2018playing}
Yusuf Aytar, Tobias Pfaff, David Budden, Tom~Le Paine, Ziyu Wang, and Nando
  de~Freitas.
\newblock Playing hard exploration games by watching youtube.
\newblock \emph{arXiv preprint arXiv:1805.11592}, 2018.

\bibitem[Bagnell et~al.(2004)Bagnell, Kakade, Schneider, and
  Ng]{bagnell2004policy}
J~Andrew Bagnell, Sham~M Kakade, Jeff~G Schneider, and Andrew~Y Ng.
\newblock Policy search by dynamic programming.
\newblock In \emph{Advances in neural information processing systems}, pages
  831--838, 2004.

\bibitem[Bellemare et~al.(2016)Bellemare, Srinivasan, Ostrovski, Schaul,
  Saxton, and Munos]{bellemare2016unifying}
Marc Bellemare, Sriram Srinivasan, Georg Ostrovski, Tom Schaul, David Saxton,
  and Remi Munos.
\newblock Unifying count-based exploration and intrinsic motivation.
\newblock In \emph{Advances in Neural Information Processing Systems}, pages
  1471--1479, 2016.

\bibitem[Bellemare et~al.(2013)Bellemare, Naddaf, Veness, and
  Bowling]{bellemare2013arcade}
Marc~G Bellemare, Yavar Naddaf, Joel Veness, and Michael Bowling.
\newblock The arcade learning environment: An evaluation platform for general
  agents.
\newblock \emph{Journal of Artificial Intelligence Research}, 47:\penalty0
  253--279, 2013.

\bibitem[Chen et~al.(2017)Chen, Schulman, Abbeel, and Sidor]{chen2017ucb}
Richard~Y Chen, John Schulman, Pieter Abbeel, and Szymon Sidor.
\newblock {UCB} exploration via {Q}-ensembles.
\newblock \emph{arXiv preprint arXiv:1706.01502}, 2017.

\bibitem[Dearden et~al.(1999)Dearden, Friedman, and Andre]{dearden1999model}
Richard Dearden, Nir Friedman, and David Andre.
\newblock Model based bayesian exploration.
\newblock In \emph{Proceedings of the Fifteenth conference on Uncertainty in
  artificial intelligence}, pages 150--159. Morgan Kaufmann Publishers Inc.,
  1999.

\bibitem[Espeholt et~al.(2018)Espeholt, Soyer, Munos, Simonyan, Mnih, Ward,
  Doron, Firoiu, Harley, Dunning, et~al.]{espeholt2018impala}
Lasse Espeholt, Hubert Soyer, Remi Munos, Karen Simonyan, Volodymir Mnih, Tom
  Ward, Yotam Doron, Vlad Firoiu, Tim Harley, Iain Dunning, et~al.
\newblock Impala: Scalable distributed deep-rl with importance weighted
  actor-learner architectures.
\newblock \emph{arXiv preprint arXiv:1802.01561}, 2018.

\bibitem[Florensa et~al.(2017)Florensa, Held, Wulfmeier, and
  Abbeel]{florensa2017reverse}
Carlos Florensa, David Held, Markus Wulfmeier, and Pieter Abbeel.
\newblock Reverse curriculum generation for reinforcement learning.
\newblock \emph{arXiv preprint arXiv:1707.05300}, 2017.

\bibitem[Garmulewicz et~al.(2018)Garmulewicz, Michalewski, and
  Mi{\l}o{\'s}]{garmulewicz2018expert}
Micha{\l} Garmulewicz, Henryk Michalewski, and Piotr Mi{\l}o{\'s}.
\newblock Expert-augmented actor-critic for vizdoom and montezumas revenge.
\newblock \emph{arXiv preprint arXiv:1809.03447}, 2018.

\bibitem[Haarnoja et~al.(2017)Haarnoja, Tang, Abbeel, and
  Levine]{haarnoja2017reinforcement}
Tuomas Haarnoja, Haoran Tang, Pieter Abbeel, and Sergey Levine.
\newblock Reinforcement learning with deep energy-based policies.
\newblock \emph{arXiv preprint arXiv:1702.08165}, 2017.

\bibitem[Hester et~al.(2017)Hester, Vecerik, Pietquin, Lanctot, Schaul, Piot,
  Horgan, Quan, Sendonaris, Dulac-Arnold, et~al.]{hester2017deep}
Todd Hester, Matej Vecerik, Olivier Pietquin, Marc Lanctot, Tom Schaul, Bilal
  Piot, Dan Horgan, John Quan, Andrew Sendonaris, Gabriel Dulac-Arnold, et~al.
\newblock Deep q-learning from demonstrations.
\newblock \emph{arXiv preprint arXiv:1704.03732}, 2017.

\bibitem[Hosu and Rebedea(2016)]{hosu2016playing}
Ionel-Alexandru Hosu and Traian Rebedea.
\newblock Playing atari games with deep reinforcement learning and human
  checkpoint replay.
\newblock \emph{arXiv preprint arXiv:1607.05077}, 2016.

\bibitem[Kakade(2002)]{kakade2002natural}
Sham~M Kakade.
\newblock A natural policy gradient.
\newblock In \emph{Advances in neural information processing systems}, pages
  1531--1538, 2002.

\bibitem[Kolter and Ng(2009)]{kolter2009near}
J~Zico Kolter and Andrew~Y Ng.
\newblock Near-bayesian exploration in polynomial time.
\newblock In \emph{Proceedings of the 26th Annual International Conference on
  Machine Learning}, pages 513--520. ACM, 2009.

\bibitem[Machado et~al.(2017)Machado, Bellemare, Talvitie, Veness, Hausknecht,
  and Bowling]{machado2017revisiting}
Marlos~C Machado, Marc~G Bellemare, Erik Talvitie, Joel Veness, Matthew
  Hausknecht, and Michael Bowling.
\newblock Revisiting the arcade learning environment: Evaluation protocols and
  open problems for general agents.
\newblock \emph{arXiv preprint arXiv:1709.06009}, 2017.

\bibitem[Mnih et~al.(2015)Mnih, Kavukcuoglu, Silver, Rusu, Veness, Bellemare,
  Graves, Riedmiller, Fidjeland, Ostrovski, et~al.]{mnih2015human}
Volodymyr Mnih, Koray Kavukcuoglu, David Silver, Andrei~A Rusu, Joel Veness,
  Marc~G Bellemare, Alex Graves, Martin Riedmiller, Andreas~K Fidjeland, Georg
  Ostrovski, et~al.
\newblock Human-level control through deep reinforcement learning.
\newblock \emph{Nature}, 518\penalty0 (7540):\penalty0 529, 2015.

\bibitem[Mnih et~al.(2016)Mnih, Badia, Mirza, Graves, Lillicrap, Harley,
  Silver, and Kavukcuoglu]{mnih2016asynchronous}
Volodymyr Mnih, Adria~Puigdomenech Badia, Mehdi Mirza, Alex Graves, Timothy~P
  Lillicrap, Tim Harley, David Silver, and Koray Kavukcuoglu.
\newblock Asynchronous methods for deep reinforcement learning.
\newblock In \emph{International Conference on Machine Learning}, 2016.

\bibitem[Nachum et~al.(2017)Nachum, Norouzi, Xu, and
  Schuurmans]{nachum2017bridging}
Ofir Nachum, Mohammad Norouzi, Kelvin Xu, and Dale Schuurmans.
\newblock Bridging the gap between value and policy based reinforcement
  learning.
\newblock In \emph{Advances in Neural Information Processing Systems}, pages
  2772--2782, 2017.

\bibitem[Nair et~al.(2018)Nair, McGrew, Andrychowicz, Zaremba, and
  Abbeel]{nair2018overcoming}
Ashvin Nair, Bob McGrew, Marcin Andrychowicz, Wojciech Zaremba, and Pieter
  Abbeel.
\newblock Overcoming exploration in reinforcement learning with demonstrations.
\newblock In \emph{2018 IEEE International Conference on Robotics and
  Automation (ICRA)}, pages 6292--6299. IEEE, 2018.

\bibitem[O'Donoghue et~al.(2016)O'Donoghue, Munos, Kavukcuoglu, and
  Mnih]{o2016pgq}
Brendan O'Donoghue, Remi Munos, Koray Kavukcuoglu, and Volodymyr Mnih.
\newblock Pgq: Combining policy gradient and q-learning.
\newblock \emph{arXiv preprint arXiv:1611.01626}, 2016.

\bibitem[Ostrovski et~al.(2017)Ostrovski, Bellemare, Oord, and
  Munos]{ostrovski2017count}
Georg Ostrovski, Marc~G Bellemare, Aaron van~den Oord, and R{\'e}mi Munos.
\newblock Count-based exploration with neural density models.
\newblock \emph{arXiv preprint arXiv:1703.01310}, 2017.

\bibitem[Pathak et~al.(2017)Pathak, Agrawal, Efros, and
  Darrell]{pathak2017curiosity}
Deepak Pathak, Pulkit Agrawal, Alexei~A Efros, and Trevor Darrell.
\newblock Curiosity-driven exploration by self-supervised prediction.
\newblock In \emph{International Conference on Machine Learning (ICML)}, volume
  2017, 2017.

\bibitem[Peng et~al.(2018)Peng, Abbeel, Levine, and van~de
  Panne]{peng2018deepmimic}
Xue~Bin Peng, Pieter Abbeel, Sergey Levine, and Michiel van~de Panne.
\newblock Deepmimic: Example-guided deep reinforcement learning of
  physics-based character skills.
\newblock \emph{arXiv preprint arXiv:1804.02717}, 2018.

\bibitem[Pohlen et~al.(2018)Pohlen, Piot, Hester, Azar, Horgan, Budden,
  Barth-Maron, van Hasselt, Quan, Ve{\v{c}}er{\'\i}k,
  et~al.]{pohlen2018observe}
Tobias Pohlen, Bilal Piot, Todd Hester, Mohammad~Gheshlaghi Azar, Dan Horgan,
  David Budden, Gabriel Barth-Maron, Hado van Hasselt, John Quan, Mel
  Ve{\v{c}}er{\'\i}k, et~al.
\newblock Observe and look further: Achieving consistent performance on atari.
\newblock \emph{arXiv preprint arXiv:1805.11593}, 2018.

\bibitem[Resnick et~al.(2018)Resnick, Raileanu, Kapoor, Peysakhovich, Cho, and
  Bruna]{resnick2018backplay}
Cinjon Resnick, Roberta Raileanu, Sanyam Kapoor, Alex Peysakhovich, Kyunghyun
  Cho, and Joan Bruna.
\newblock Backplay:" man muss immer umkehren".
\newblock \emph{arXiv preprint arXiv:1807.06919}, 2018.

\bibitem[Schaul et~al.(2015)Schaul, Quan, Antonoglou, and
  Silver]{schaul2015prioritized}
Tom Schaul, John Quan, Ioannis Antonoglou, and David Silver.
\newblock Prioritized experience replay.
\newblock \emph{arXiv preprint arXiv:1511.05952}, 2015.

\bibitem[Schulman et~al.(2017{\natexlab{a}})Schulman, Abbeel, and
  Chen]{schulman2017equivalence}
John Schulman, Pieter Abbeel, and Xi~Chen.
\newblock Equivalence between policy gradients and soft q-learning.
\newblock \emph{arXiv preprint arXiv:1704.06440}, 2017{\natexlab{a}}.

\bibitem[Schulman et~al.(2017{\natexlab{b}})Schulman, Wolski, Dhariwal,
  Radford, and Klimov]{schulman2017proximal}
John Schulman, Filip Wolski, Prafulla Dhariwal, Alec Radford, and Oleg Klimov.
\newblock Proximal policy optimization algorithms.
\newblock \emph{arXiv preprint arXiv:1707.06347}, 2017{\natexlab{b}}.

\bibitem[Sutton et~al.(2000)Sutton, McAllester, Singh, and
  Mansour]{sutton2000policy}
Richard~S Sutton, David~A McAllester, Satinder~P Singh, and Yishay Mansour.
\newblock Policy gradient methods for reinforcement learning with function
  approximation.
\newblock In \emph{Advances in neural information processing systems}, pages
  1057--1063, 2000.

\bibitem[Tang et~al.(2017)Tang, Houthooft, Foote, Stooke, Chen, Duan, Schulman,
  De~Turck, and Abbeel]{Tang2016exploration}
Haoran Tang, Rein Houthooft, Davis Foote, Adam Stooke, Xi~Chen, Yan Duan, John
  Schulman, Filip De~Turck, and Pieter Abbeel.
\newblock \#{E}xploration: A study of count-based exploration for deep
  reinforcement learning.
\newblock \emph{Advances in Neural Information Processing Systems (NIPS)},
  2017.

\bibitem[Van Den~Oord et~al.(2016)Van Den~Oord, Dieleman, Zen, Simonyan,
  Vinyals, Graves, Kalchbrenner, Senior, and Kavukcuoglu]{van2016wavenet}
A{\"a}ron Van Den~Oord, Sander Dieleman, Heiga Zen, Karen Simonyan, Oriol
  Vinyals, Alex Graves, Nal Kalchbrenner, Andrew~W Senior, and Koray
  Kavukcuoglu.
\newblock Wavenet: A generative model for raw audio.
\newblock In \emph{SSW}, page 125, 2016.

\bibitem[Watkins and Dayan(1992)]{watkins1992q}
Christopher~JCH Watkins and Peter Dayan.
\newblock Q-learning.
\newblock \emph{Machine learning}, 8\penalty0 (3-4):\penalty0 279--292, 1992.

\bibitem[Williams(1992)]{williams1992simple}
Ronald~J Williams.
\newblock Simple statistical gradient-following algorithms for connectionist
  reinforcement learning.
\newblock \emph{Machine learning}, 8\penalty0 (3-4):\penalty0 229--256, 1992.

\bibitem[Wu et~al.(2017)Wu, Mansimov, Grosse, Liao, and Ba]{wu2017scalable}
Yuhuai Wu, Elman Mansimov, Roger~B Grosse, Shun Liao, and Jimmy Ba.
\newblock Scalable trust-region method for deep reinforcement learning using
  kronecker-factored approximation.
\newblock In \emph{Advances in neural information processing systems}, pages
  5279--5288, 2017.

\end{thebibliography}
\bibliographystyle{plainnat}

\end{document}